\PassOptionsToPackage{table}{xcolor}
\PassOptionsToPackage{capitalize,noabbrev}{cleveref}

\documentclass[]{preprint}

\usepackage[toc,page,header]{appendix}

\usepackage[T1]{fontenc}
\usepackage{listings}
\usepackage{tcolorbox}
\tcbuselibrary{listings,breakable}
\usepackage{xcolor}
\usepackage{upquote}
\usepackage{circledsteps}

\usepackage{microtype}
\usepackage{graphicx}
\usepackage{subcaption}
\usepackage{booktabs} 

\usepackage{hyperref}

\usepackage{amsmath}
\usepackage{amssymb}
\usepackage{mathtools}
\usepackage{amsthm}

\usepackage{stmaryrd}

\theoremstyle{plain}

\theoremstyle{definition}

\theoremstyle{remark}

\usepackage{hyperref}
\usepackage{url}
\usepackage{xcolor}
\usepackage{pifont}

\usepackage[utf8]{inputenc}
\usepackage{csquotes} 
\usepackage{enumitem}
\usepackage{multirow}
\usepackage{array}
\usepackage{colortbl}

\usepackage{wrapfig}
\usepackage{subcaption}  
\usepackage{dblfloatfix}

\usepackage[most]{tcolorbox}
\usepackage{xcolor}
\usepackage{graphicx}

\definecolor{cvprblue}{RGB}{0,102,204} 

\usepackage{amsmath,bm}

\usepackage{color}
\usepackage{tikz}
\usetikzlibrary{shapes, arrows.meta, positioning, calc, fit, backgrounds}

\usepackage{etoc}

\definecolor{sybcblue}{HTML}{004488}  
\definecolor{brickred}{HTML}{BB5566}  
\definecolor{softgray}{HTML}{666666}  

\definecolor{graybg}{gray}{0.9}

\usepackage{hyperref}
\hypersetup{
    colorlinks=true,
    linkcolor=black,
    filecolor=magenta,      
    urlcolor=seedblue, 
    pdftitle={Paper Title},
}

\usepackage{tikz}

\usepackage{cleveref}

\definecolor{softred}{RGB}{255,230,230}
\definecolor{softblue}{RGB}{230,240,255}
\definecolor{slowcolor}{RGB}{0,100,200} 

\usepackage{amsmath}   
\usepackage{amssymb}   
\usepackage{amsthm}    

\usepackage{tcolorbox}

\usepackage{enumitem}

\usepackage{amsmath,amssymb,amsthm,mathtools}

\usepackage[utf8]{inputenc} 
\usepackage[T1]{fontenc}    
\usepackage{hyperref}       
\usepackage{url}            
\usepackage{booktabs}       
\usepackage{amsfonts}       
\usepackage{nicefrac}       
\usepackage{microtype}      
\usepackage{xcolor}         

\usepackage[utf8]{inputenc} 
\usepackage[T1]{fontenc}    
\usepackage{hyperref}       
\usepackage{url}            
\usepackage{booktabs}       
\usepackage{amsfonts}       
\usepackage{nicefrac}       
\usepackage{microtype}      
\usepackage{xcolor}         

\usepackage{tikz}
\usetikzlibrary{arrows.meta, calc, shapes.geometric, decorations.pathreplacing, positioning}

\usepackage[utf8]{inputenc} 
\usepackage[T1]{fontenc}    
\usepackage{url}            
\usepackage{booktabs}       
\usepackage{amsfonts}       
\usepackage{nicefrac}       
\usepackage{microtype}      
\usepackage{xcolor}         

\usepackage{amsmath}
\usepackage{amssymb}

\usepackage[utf8]{inputenc} 
\usepackage[T1]{fontenc}    
\usepackage{url}            
\usepackage{booktabs}       
\usepackage{amsfonts}       
\usepackage{nicefrac}       
\usepackage{microtype}      

\usepackage{wrapfig}

\usepackage{pifont}

\usepackage[table]{xcolor}
\usepackage{booktabs}
\usepackage{multirow}

\usepackage{booktabs}
\usepackage{tabularx}

\usepackage{tikz}
\usetikzlibrary{spy, arrows.meta, calc, decorations.pathreplacing, bending, shadows} 

\usepackage{amsmath,amssymb}
\usepackage{tikz}
\usetikzlibrary{
  arrows.meta,
  positioning,
  calc,
  fit,
  backgrounds,
  decorations.pathreplacing,
  shapes.geometric,
  shapes.misc,
  matrix
}

\usepackage{tikz}

\usepackage{xcolor}

\usetikzlibrary{
    arrows.meta,      
    calc,             
    shapes.geometric, 
    fadings,          
    bending           
}

\usepackage{tikz}
\usetikzlibrary{arrows.meta, calc, decorations.pathreplacing, positioning}
\usepackage{amsmath, amssymb}

\usepackage{tikz}
\usetikzlibrary{arrows.meta, calc, angles, quotes}

\usepackage{booktabs}
\usepackage{multirow}
\usepackage{graphicx}
\usepackage[table]{xcolor}
\usepackage{xspace}


\definecolor{graybg}{gray}{0.92}

\definecolor{promptblue}{HTML}{4F78B6}

\tcbset{
  promptbox/.style={
    enhanced,
    breakable,
    colframe=promptblue,
    colback=promptblue!3,
    opacityback=0.95,
    boxrule=0.7pt,
    arc=4pt,
    left=6pt,
    right=6pt,
    top=5pt,
    bottom=5pt,
    fonttitle=\bfseries,
    coltitle=white,
    colbacktitle=promptblue,
    boxed title style={
      arc=3pt,
      boxrule=0pt,
      left=5pt,
      right=5pt,
      top=2pt,
      bottom=2pt
    },
    attach boxed title to top left={
      xshift=6pt,
      yshift=-2pt
    }
  }
}

\usepackage{tikz}
\usetikzlibrary{arrows.meta, calc}
\usepackage{amsmath, amssymb}

\lstdefinestyle{prompt}{
    basicstyle=\ttfamily\scriptsize,
    breaklines=true,
    columns=fullflexible,
    frame=single,
    framerule=0.3pt,
    xleftmargin=0.6em,
    xrightmargin=0.6em,
    aboveskip=0.5em,
    belowskip=0.5em,
    showstringspaces=false
}
\usepackage{enumitem}
\usepackage{listings}
\usepackage{tcolorbox}
\tcbuselibrary{skins,breakable}
\definecolor{LightGray}{HTML}{BFBFBF}
\newtcolorbox{promptbox}[2][]{%
    enhanced,
    breakable,
    colback=gray!3,
    colframe=black!45,
    colbacktitle=gray!12,
    coltitle=black,
    fonttitle=\bfseries,
    title={#2},
    boxrule=0.5pt,
    arc=2pt,
    left=8pt,
    right=8pt,
    top=8pt,
    bottom=8pt,
    before skip=10pt,
    after skip=10pt,
    #1
}

\lstdefinestyle{prompt}{%
    basicstyle=\ttfamily\footnotesize,
    columns=fullflexible,
    keepspaces=true,
    breaklines=true,
    breakatwhitespace=false,
    showstringspaces=false,
    numbers=none,
    frame=none,
    aboveskip=6pt,
    belowskip=2pt
}

\definecolor{src}{HTML}{1B4965}
\definecolor{tgt}{HTML}{C44536}
\definecolor{gold}{HTML}{B8860B}
\definecolor{ell}{HTML}{C5D8E8}
\definecolor{gridg}{HTML}{DCDCDC}
\definecolor{note}{HTML}{6E7F8E}

\definecolor{CaseBlue}{HTML}{2471A3}

\usepackage{custom} 

\usepackage{adjustbox}
\definecolor{mygreen}{RGB}{70,170,80}
\definecolor{myred}{RGB}{200,30,30}
\newcommand{\graytext}[1]{\textcolor{gray}{#1}}
\definecolor{revisoblue}{RGB}{226,238,248}
\definecolor{revisopink}{RGB}{249,230,234}
\definecolor{revisoyellow}{RGB}{250,242,214}
\definecolor{revisogreen}{RGB}{220,245,220}

\hypersetup{pdftitle={Reviso: Discovering What Must Be Seen for Agentic Video Captioning}}

\newlength{\smallfloatwidth}
\usepackage{float}
\makeatletter
\newcommand{\wrapneedspace}[1]{%
  \par\penalty-100\begingroup
  \dimen@=\pagegoal\advance\dimen@-\pagetotal
  \ifdim\dimen@<#1\relax\newpage\fi
  \endgroup
}
\newcommand{\finishwrap}{%
  \par
  \ifnum\c@WF@wrappedlines>1
    \begingroup
    \@tempcnta=\c@WF@wrappedlines\advance\@tempcnta by -1
    \dimen@=\baselineskip\multiply\dimen@ by \@tempcnta
    \vspace{\dimen@}
    \endgroup
  \fi
  \WFclear
}
\makeatother

\title{Reviso: Not All Visual Evidence Deserves Equal Attention in Video Captioning}
\author[1]{Tengfei~Liu}
\author[1,2,\dagger,\spadesuit]{Yang~Shi}
\author[1]{Yuran~Wang}
\author[3]{Yuqi~Tang}
\author[4]{Yizhuo~Jia}
\author[1]{Zhuoran~Zhang}
\author[1]{Qixun~Wang}
\author[5]{Shujie~Li}
\author[1]{Xuanyu~Zhu}
\author[6]{Xinyu~Liu}
\author[7]{Minghui~Guo}
\author[8,\dagger]{Shixuan~Liu}
\author[8,\dagger]{Haotian~Wang}
\author[2,\spadesuit]{Yuanxing~Zhang}
\author[2]{Pengfei~Wan}
\affiliation[1]{PKU}
\affiliation[2]{KlingAI Research}
\affiliation[3]{HKUST(GZ)}
\affiliation[4]{FDU}
\affiliation[5]{HKU}
\affiliation[6]{HKUST}
\affiliation[7]{NUS}
\affiliation[8]{THU}
\providecommand{\firstpagefootnotes}{}
\renewcommand{\firstpagefootnotes}{%
  \parbox{0.96\textwidth}{%
    {\color{seedblue}\hrule height 0.2pt\relax}
    \vspace{6pt}
    \raggedright\normalsize
    $^{\dagger}$ Corresponding Author. \qquad $^{\spadesuit}$ Project Lead.
  }%

}

\abstract{Multimodal large language models (MLLMs) can generate detailed video captions, yet they may overemphasize minor details while overlooking essential context and key events.
We argue that not all visual evidence deserves equal attention, as the importance of visual information depends on the content of each video.
Motivated by this observation, we propose \textbf{Reviso}, an agentic video captioning framework that learns what to observe and how to gather relevant visual evidence.
From a low-frame-rate global preview, Reviso constructs a hierarchical Context--Focus--Points (CFP) observation plan that identifies the video context, prioritizes key aspects, and defines concrete observation targets and corresponding tools.
Guided by this plan, Reviso performs goal-directed active perception by selectively revisiting temporal segments and combining tools for fine-grained inspection.
It interprets the acquired evidence, updates its plan as needed, and integrates global and local information to generate the final caption.
To learn this observation strategy, we develop a verification-guided pipeline that constructs high-quality training trajectories through factual verification and teacher-guided refinement.
We further train Reviso using Group Relative Policy Optimization (GRPO), incorporating a key-fact reward for caption quality, a Plan Alignment Reward (PAR) for identifying important observation targets, and an Unproductive Tool-Call Penalty (UTCP) that adjusts observation-cost penalties based on caption quality.
Experiments across four video captioning benchmarks show that Reviso outperforms the evaluated open-source baselines, while its generated captions better support downstream video question answering.
These findings highlight the effectiveness and potential of priority-guided active observation for video captioning.}
\date{%
\today\hfill
\href{https://github.com/pkucs-Ltf/Reviso}{\textbf{Code}}%
}
\begin{document}
\raggedbottom
\maketitle

\begin{figure}[t]
  \centering
  \includegraphics[width=0.99\linewidth]{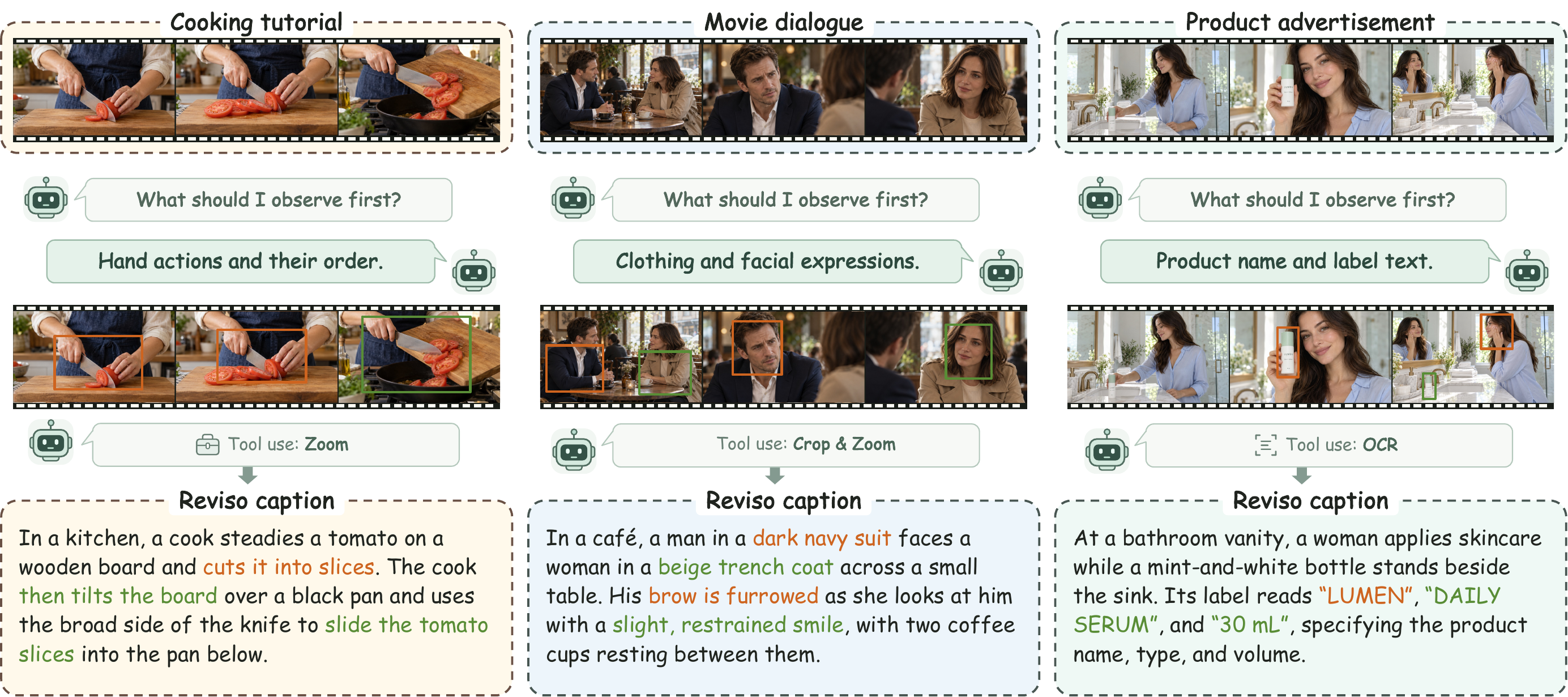}
  \caption{\textbf{Video captioning through active observation.} Captioning priorities vary across video types, including action order in tutorials, characters’ expressions and actions in movie scenes, and product names and attributes in advertisements. Reviso identifies what matters in each video and selects tools to gather evidence for its caption.}
  \label{fig:teaser}
\end{figure}

\section{Introduction}
\label{sec:intro}

Video captioning aims to capture and convey visual information from videos, playing an important role in video understanding and generation~\cite{wang2024internvideo2,yang2025cogvideox,fan2025instancecap,tang2026artifact,liu2026refcaptioner,zhang2026verus,liu2026longav,zhang2026multiref,tang2026keyframe}.
Recent advances in multimodal large language models (MLLMs)~\cite{qwen2026qwen35,qwen2026qwen38,wang2026beacon,wang2026monet,deepmind2026gemini31pro,su2026omnipack,zhang2025debiasing,openai2026gpt6astra} have enabled increasingly detailed video captions~\cite{chai2024auroracap,yang2024vript,shi2025mavors}.
However, greater descriptive detail does not necessarily ensure coverage of key information.
Existing models may overemphasize minor details while overlooking essential context and key events, particularly in videos involving complex action sequences, subtle interactions, or fine-grained visual details~\cite{wang2024tarsier,omnifysicscaptioner2026}.

We argue that \textbf{not all visual evidence deserves equal attention}.
As illustrated in Figure~\ref{fig:teaser}, the importance of visual information depends on the content and context of each video.
Moreover, important information may not be readily resolved from a coarse visual overview and may require targeted inspection~\cite{wu2024vstar,omnifysicscaptioner2026}.
Learning such an observation strategy presents two main challenges.
\textbf{(1) Identifying what to observe.} 
The model must determine which information is important for understanding the video and translate these priorities into concrete observation targets.
\textbf{(2) Learning how to observe.} 
Different observation targets may require different tools for fine-grained visual inspection~\cite{videozoomer2025,wu2024vstar,omnicaptioner2025}.
The model must select and combine appropriate tools, interpret the acquired evidence, and adapt subsequent observations accordingly.

To address these challenges, we propose \textbf{Reviso}, an agentic video captioning framework that learns to identify important visual information and actively gather supporting evidence.
As shown in Figure~\ref{fig:architecture}, Reviso first constructs a hierarchical Context-Focus-Points (CFP) observation plan from a low-frame-rate global preview.
The Context level identifies the overall video setting, Focus prioritizes important aspects, and Points define concrete observation targets and the tools needed to inspect them.
Guided by this plan, Reviso performs goal-directed active perception by selectively revisiting temporal segments at higher frame rates and using tools such as image cropping and OCR for fine-grained inspection.
Based on the acquired evidence, it determines which targets have been resolved and updates its plan to guide subsequent observations.
Finally, Reviso integrates global context and local evidence to generate the caption.

To learn this observation strategy, we develop a verification-guided pipeline that constructs high-quality agentic trajectories through factual verification and teacher-guided refinement.
These trajectories provide supervised fine-tuning data for learning observation planning, tool use, and evidence-grounded caption generation.
We further train Reviso using GRPO~\cite{shao2024deepseekmath}, incorporating a Key-Fact Reward (KFR) for caption quality, a Plan Alignment Reward (PAR) for identifying important observation targets, and an Unproductive Tool-Call Penalty (UTCP) to regulate observation costs based on caption quality.
Extensive experiments across four video captioning benchmarks show that Reviso consistently improves over its backbone and outperforms the strong open-source baselines. 
Further evaluations demonstrate that Reviso generates captions that better support downstream video understanding, highlighting the effectiveness of learning to prioritize and actively gather relevant visual evidence.

Our contributions are summarized as follows:

\begin{itemize}
    \item We introduce a \textbf{priority-aware perspective on video captioning}, arguing that not all visual evidence deserves equal attention. This perspective emphasizes identifying video-specific observation priorities and actively gathering evidence for key information, rather than treating all visual content equally.
    \item We propose \textbf{Reviso}, an agentic video captioning framework that realizes this perspective through hierarchical Context-Focus-Points (CFP) planning and goal-directed active perception. We further develop a verification-guided trajectory construction pipeline and tailored GRPO rewards to learn effective observation planning and tool use.
    \item Extensive experiments across four video captioning benchmarks demonstrate that Reviso improves caption quality over strong open-source baselines. Further evaluations show that its generated captions better support downstream video understanding.
\end{itemize}

\section{Related Work}
\label{sec:related_work}

\textbf{Video Captioning.}
Recent video captioning methods~\cite{liu2026refcaptioner,xue2025progresscaptioner,wei2025longcaptioning,chai2024auroracap} have explored various approaches to capturing fine-grained visual information and temporal dynamics.
Tarsier~\cite{wang2024tarsier}improves detailed video description through large-scale pretraining and instruction tuning. TimeChat-Captioner~\cite{yao2026timechatcaptioner} generates timestamped, structured audio-visual descriptions, while PercepCap~\cite{xu2026percepcap} produces a structured record of object
trajectories and temporal events before generating the caption. Reinforcement learning further improves caption quality. CapRL++~\cite{yang2026caprlpp} rewards captions based on whether a text-only model can correctly answer video-related questions using only those captions, while CuRe~\cite{gao2026cure} uses claim-level verification and
reference-anchored calibration to reward caption's factual accuracy and coverage. For long videos, PIGEONS~\cite{chen2026pigeons} maintains a persistent identity gallery and an evolving narrative summary to preserve subject consistency across clips.
These approaches primarily improve caption generation from available visual representations, without explicitly learning to identify video-specific observation targets and acquire additional evidence to resolve them.
In contrast, Reviso learns to prioritize important information and actively gather supporting visual evidence through goal-directed observation.

\textbf{Agentic Video Understanding.}
Video agents combine reasoning with iterative observation to acquire relevant visual evidence.
VideoAgent~\cite{wang2024videoagent} retrieves additional frames, while VideoTree~\cite{wang2024videotree} constructs query-adaptive hierarchical video representations.
Recent methods further explore learned observation policies.
VITAL~\cite{zhang2025vital}, VideoZoomer~\cite{videozoomer2025}, and VideoTemp-o3~\cite{liu2026videotempo3} selectively revisit temporal segments, while EVA~\cite{zhang2026eva} learns to adjust observation intervals and sampling configurations.
Other approaches explore multi-agent collaboration~\cite{videochatm12025} and integrate video inspection with external search~\cite{videorover2026}.
Many of these methods focus on query-driven video understanding, where observation is guided by a given question or instruction.
In contrast, open-ended video captioning requires models to identify important information and determine what evidence to gather without an explicit question.
Omni-Detective~\cite{omnicaptioner2025} and OmniFysics-Agent~\cite{omnifysicscaptioner2026} employ agents to collect evidence for caption generation, but their final captioners learn primarily from the resulting descriptions rather than the observation trajectories.
Reviso instead learns video-specific observation planning and tool-assisted evidence gathering through verified trajectory supervision and reinforcement learning.

\section{Method}
\label{sec:method}

\subsection{Priority-Guided Active Observation}
\label{sec:observation}

\begin{figure}[htbp]
\centering
\includegraphics[width=\textwidth]{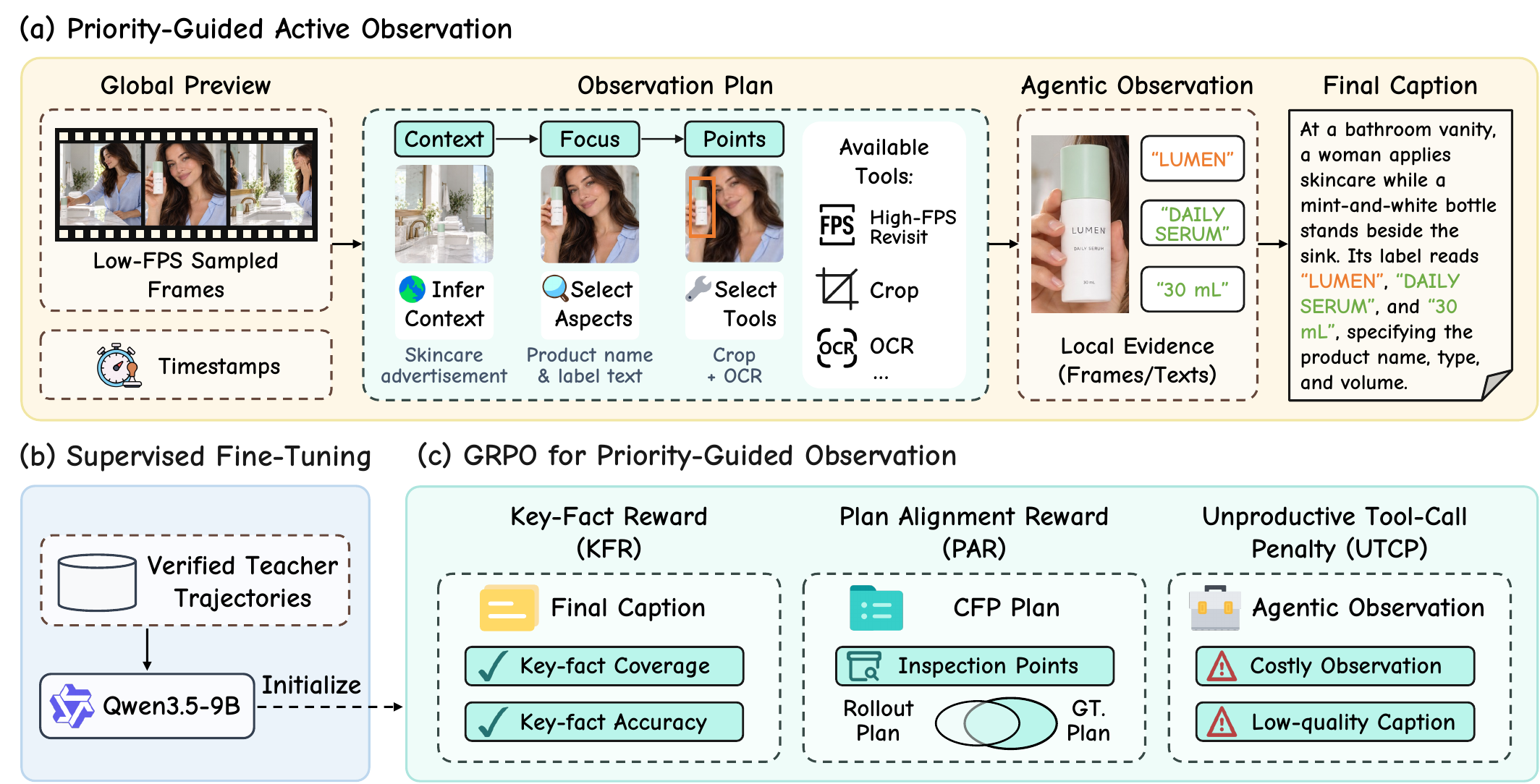}
\caption{\textbf{Overview of Reviso.} (a) A hierarchical Context–Focus–Points plan guides tool-assisted observation and caption generation.
(b) Verified teacher trajectories provide supervised initialization. (c) GRPO rewards factual caption quality (KFR) and
plan alignment (PAR), while penalizing unproductive tool use (UTCP).}
\label{fig:architecture}
\end{figure}

\noindent\textbf{Global preview and planning.}
As shown in Figure~\ref{fig:architecture}(a), Reviso first samples timestamped frames at a low frame rate to obtain a global overview of the video.
Based on this preview, it constructs a hierarchical Context--Focus--Points (CFP) observation plan that identifies video-specific priorities and translates them into concrete observation targets and corresponding tool actions.

\begin{figure}[htbp]
\centering
\includegraphics[width=\textwidth]{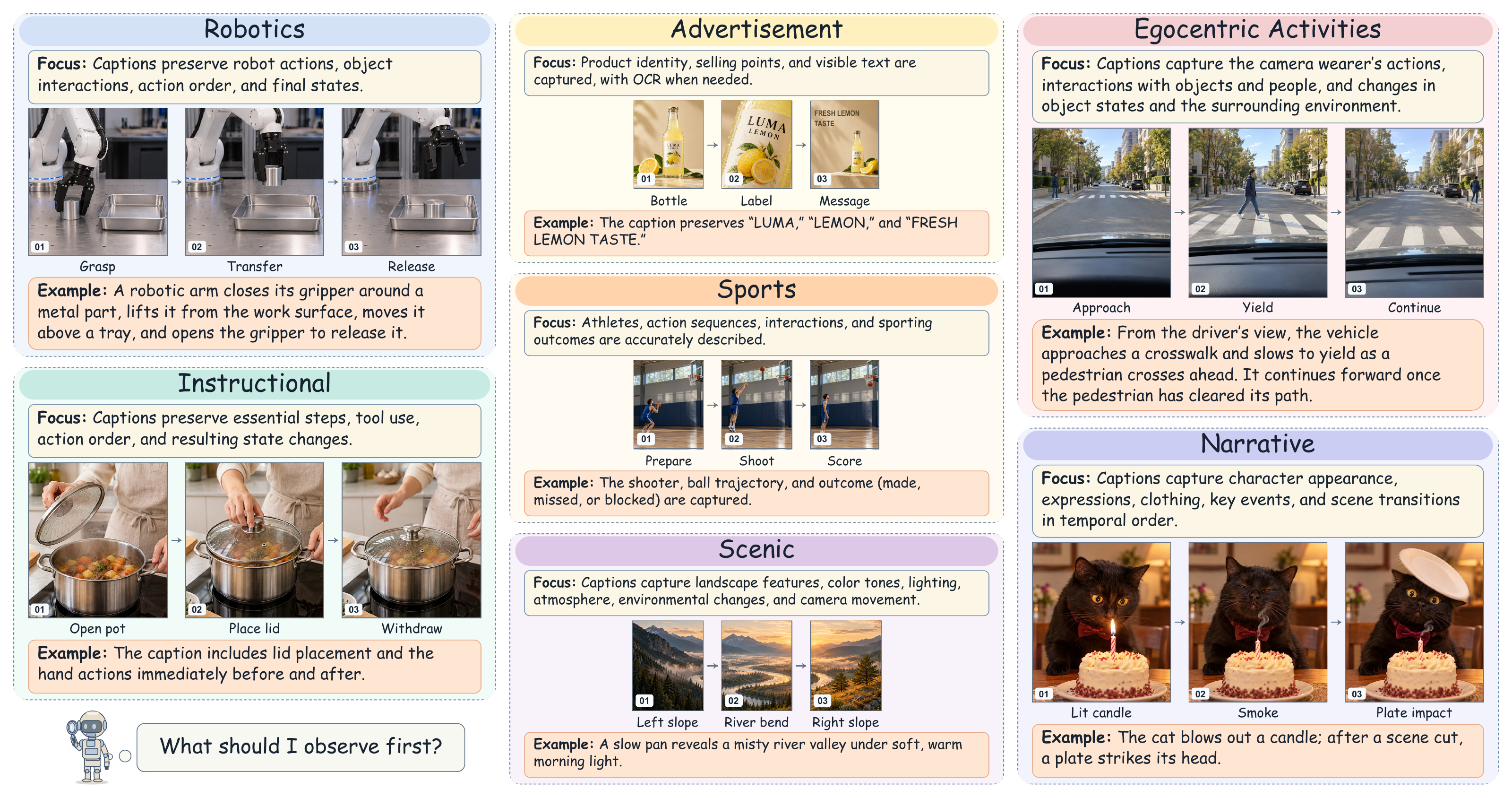}
\caption{\textbf{Observation priorities across diverse video content.} We illustrate representative visual aspects that may require prioritized attention in different video scenarios. Reviso identifies important information based on the content and context of each video, guiding targeted observation and evidence gathering.}
\label{fig:class}
\end{figure}

\begin{itemize}
\item\indent\textbf{Context:}
Reviso identifies the overall context of the video, including its setting, main activities, and relevant background information.
This context provides the basis for assessing the importance of different visual information.
\item\indent\textbf{Focus:}
Based on the inferred context, Reviso prioritizes the aspects most relevant to understanding the video.
As illustrated in Figure~\ref{fig:class}, observation priorities vary with the content and context of each video. 
These priorities determine which information deserves closer inspection and guide the formulation of concrete observation targets at the Points level.
\item\indent\textbf{Points:}
Reviso translates each focus into concrete targets to inspect or verify and specifies the corresponding observation tools.
These targets guide the selection of relevant temporal segments, observation granularity, and tool actions, such as high-frame-rate revisiting, image cropping, and OCR.
\end{itemize}

\noindent\textbf{Agentic fine-grained observation.}
Guided by the CFP plan, Reviso performs multiple rounds of tool-assisted observation to gather evidence for prioritized information.
At each round, it selects observation targets based on the current plan and invokes appropriate tools according to the evidence needed.
The resulting observations are interpreted together with the global context and previously acquired evidence.
Reviso then identifies resolved and unresolved targets and updates its plan to direct further inspection toward important information that remains insufficiently observed.
Finally, it integrates the global context with the collected local evidence to generate the final video caption.

\subsection{Training Strategy}
\label{sec:policy_learning}

\subsubsection{Supervised Fine-Tuning}
We initialize Reviso from Qwen3.5-9B~\cite{qwen2026qwen35} and fine-tune it on verified teacher trajectories, as shown in Figure~\ref{fig:architecture}(b).
Token-level supervision covers observation plans, tool calls, local interpretations, plan updates, and the final caption.
Environment responses provide context and are masked from the loss.
This stage initializes the policy for the following stage.

\subsubsection{GRPO for Priority-Guided Observation}
Following supervised fine-tuning, we further train Reviso using GRPO~\cite{shao2024deepseekmath} with three complementary reward components: Key-Fact Reward (KFR) for caption quality, Plan Alignment Reward (PAR) for observation planning, and Unproductive Tool-Call Penalty (UTCP) for regulating observation costs.

\noindent\textbf{Key-Fact Reward (KFR).}
For a video $V$, let $\tau$ denote a generated observation trajectory and $c$ its final caption.
We decompose the reference caption and generated caption into deduplicated fact sets $\mathcal F_V$ and $\widehat{\mathcal F}(c)$, respectively.
KFR is defined as
\begin{equation}
R_{\mathrm{KFR}}(\tau)=\operatorname{F_1}\bigl(\widehat{\mathcal F}(c),\mathcal F_V\bigr).
\label{eq}
\end{equation}
We compute F1 using precision and recall based on one-to-one semantic matching between the two fact sets.
Precision measures the proportion of generated facts matched to the reference, while recall measures the coverage of reference facts.
KFR thus encourages the preservation of key information while penalizing unmatched claims.

\noindent\textbf{Plan Alignment Reward (PAR).}
PAR evaluates the initial observation plan $p_0$ before tool execution.
Let $\mathcal P(p_0)$ denote the set of deduplicated observation targets extracted from $p_0$, interpreted within their associated Context and Focus levels.
Let $\mathcal U_V$ denote the reference observation targets corresponding to key information not resolved by the global preview but identifiable as requiring further inspection.
We measure alignment between predicted and reference targets using
\begin{equation}
R_{\mathrm{PAR}}(\tau)=\operatorname{F_1}\bigl(\mathcal P(p_0),\mathcal U_V\bigr).
\label{eq}
\end{equation}
As in KFR, we use one-to-one semantic matching to compute precision and recall.
PAR encourages the initial plan to identify relevant observation targets while limiting unmatched targets.

\noindent\textbf{Unproductive Tool-Call Penalty (UTCP).}
UTCP discourages costly observations that yield poor captions.
For each video, we activate the penalty only when at least one trajectory in the sampled group reaches a fixed KFR threshold.
If no trajectory reaches this threshold, we omit the cost penalty to preserve exploration.
Once activated, the penalty applies to every trajectory in the group according to its own observation cost and caption quality.
At the same cost, lower-quality captions receive a larger penalty.
At the same quality, more costly observations receive a larger penalty.
For a group $\mathcal G$ of trajectories sampled for the same video, we define
\begin{equation}
    \begin{aligned}
        g(\mathcal G)&=\mathbb I\!\left[\max_{\tau_i\in\mathcal G}R_{\mathrm{KFR}}(\tau_i)\geq\tau_q\right],\\
        P_{\mathrm{UTCP}}(\tau_i;\mathcal G)&=g(\mathcal G)\bigl(1-R_{\mathrm{KFR}}(\tau_i)\bigr)b_i,
    \end{aligned}
    \label{eq:UTCP}
\end{equation}
where $\tau_q$ is the fixed KFR threshold.
We quantify additional observation cost by the number of tool calls after the shared global preview.
We normalize this count by a fixed call budget to obtain $b_i\in[0,1]$.
The indicator $g(\mathcal G)$ determines whether the cost penalty is active for the group.

\noindent\textbf{Policy optimization.}
For each video, Reviso generates a group of trajectories containing observation plans, tool calls, and final captions.
We evaluate each trajectory using the combined reward
\begin{equation}
R(\tau_i;\mathcal G) = \alpha R_{\mathrm{KFR}}(\tau_i) + \beta R_{\mathrm{PAR}}(\tau_i) - \mu P_{\mathrm{UTCP}}(\tau_i;\mathcal G).
\label{eq:joint_reward}
\end{equation}
where $\alpha$, $\beta$, and $\mu$ control the contributions of the three components.
GRPO computes relative advantages from the rewards within each group and uses them to update the model.

\subsection{Data Construction}

\begin{figure}[t]
\centering
\includegraphics[width=\textwidth,height=0.6\textheight,keepaspectratio]{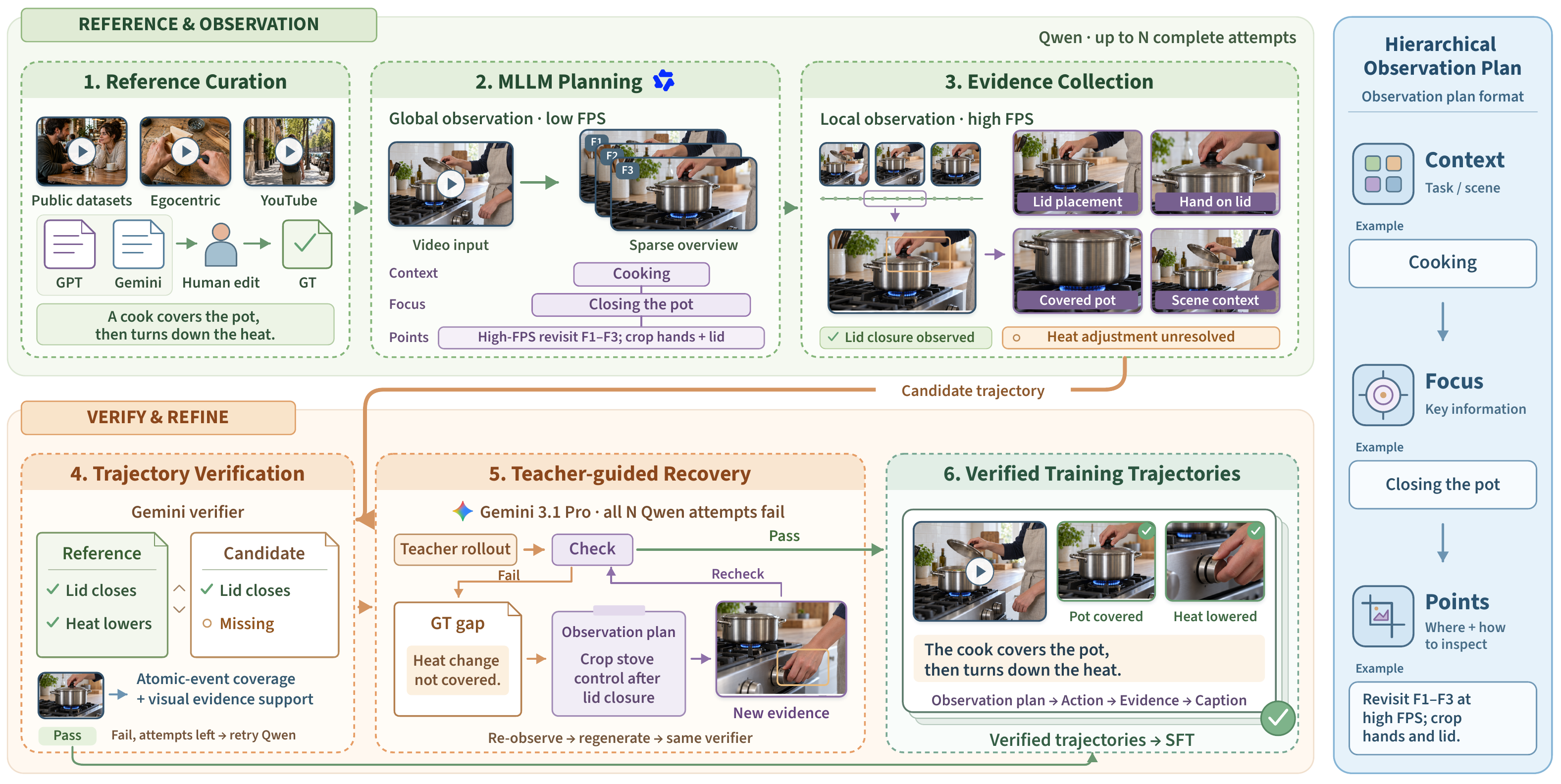}
\caption{\textbf{Overview of the trajectory construction pipeline.} There are six steps in the pipeline. We generate observation trajectories, verify them against ground-truth captions, and use a teacher model to refine unsuccessful attempts, retaining verified trajectories for supervised fine-tuning.}
\label{fig:trajectory_pipeline}
\end{figure}

As shown in Figure~\ref{fig:trajectory_pipeline}, we construct verified training trajectories through six steps.
We first collect videos and existing captions from publicly available datasets and generate additional candidate captions using GPT-6 Astra~\cite{openai2026gpt6astra} and Gemini 3.1 Pro~\cite{deepmind2026gemini31pro}.
Human annotators select, merge, correct, and supplement these descriptions to produce ground-truth captions.
Next, Qwen3.8-27B~\cite{qwen2026qwen38} examines a low-frame-rate global preview and constructs a hierarchical Context--Focus--Points (CFP) observation plan.
Following this plan, it revisits relevant temporal segments at higher frame rates and uses tools such as image cropping and OCR to inspect local details before generating a candidate caption.

For trajectory verification, we decompose the ground-truth and generated captions into atomic facts.
We use Gemini 3.1 Pro to compare the resulting fact sets for factual consistency and reference-fact coverage, and to verify whether the collected visual evidence supports the key facts in the generated caption.
If verification fails, Qwen3.8-27B repeats the complete planning, observation, and captioning process, with up to $N$ attempts per video.
Samples that fail all $N$ attempts enter teacher-guided recovery, where Gemini 3.1 Pro generates a new trajectory following the same observation workflow.
If the teacher trajectory also fails verification, the verifier provides feedback based on the ground-truth caption, identifying missing or unsupported facts.
The teacher then uses this feedback to revise its observation plan, gather additional evidence, and regenerate the caption for verification by the same verifier.
Finally, we retain verified trajectories from both Qwen3.8-27B and the teacher, including their observation plans, tool actions, observation evidence, and final captions, for supervised fine-tuning.
\section{Experiments}
\label{sec:experiments}

\subsection{Experimental Setup}
\label{sec:experimental_setup}

\textbf{Data and training.}
We initialize Reviso from Qwen3.5-9B~\cite{qwen2026qwen35}.
Our training data comprise $10,000$ trajectories for supervised fine-tuning (SFT) and $20,000$ samples for reinforcement learning (RL), drawn from LLaVA-Video-178K~\cite{zhang2024llavavideo}, Vript~\cite{yang2024vript}, Ego4D~\cite{grauman2022ego4d}, and YouTube.
The SFT and RL datasets contain no overlapping videos.
We first perform SFT with LoRA for two epochs, followed by $200$ steps of GRPO with trajectories sampled online.
The visual encoder remains frozen throughout training, while GRPO updates all other model parameters.

\noindent\textbf{Baselines.}
We compared Reviso with 15 MLLMs.
Our open-source baselines include ARC-Qwen-Video-7B~\cite{ge2025archunyuan}, Molmo2-8B~\cite{clark2026molmo2}, and Tarsier2-7B~\cite{yuan2025tarsier2}.
We also evaluate InternVL3.5~\cite{wang2025internvl3}, LLaVA-OneVision-2~\cite{an2026llavaonevision2}, and the Qwen3-VL~\cite{bai2025qwen3vl}, Qwen3.5~\cite{qwen2026qwen35}, Qwen3.6~\cite{team2026qwen327b,team2026qwen3}, and Qwen3.8~\cite{qwen2026qwen38} families.
Qwen3.5-9B is our backbone baseline.
We include GPT-6 Astra~\cite{openai2026gpt6astra} and Gemini 3.8 Flash~\cite{deepmind2026gemini38flash} as proprietary reference models.

 \noindent\textbf{Evaluation Benchmarks.}
We evaluate Reviso on 6 benchmarks, including four video captioning benchmarks and two downstream video understanding benchmarks.
On DREAM-1K~\cite{wang2024tarsier} and MCTS-VCB~\cite{yu2025mctsvcb}, we report precision, recall, and F1 to assess factual accuracy and information coverage.
CaReBench~\cite{xu2026carebench} evaluates fine-grained descriptions of actions and objects, for which we report action F1 and object F1.
On VDC~\cite{chai2024auroracap}, we report TP score and TP accuracy for detailed captions.
We further evaluate whether the generated captions support downstream video understanding on LVBench~\cite{wang2025lvbench} and MMVU~\cite{zhao2025mmvu}.
We report accuracy on LVBench and multiple-choice, open-ended, and overall accuracy on MMVU.

\noindent\textbf{Implementation details.}
For fair comparison, we limit all models to a maximum of 128 input frames per video.
Baselines use uniformly sampled frames, while Reviso distributes the same frame budget across its global preview and subsequent temporal revisits.
This ensures a consistent maximum frame budget across all evaluated methods.

\subsection{Main Results}
\label{sec:main_results}

\begin{table}[t]
    \centering
    \caption{\textbf{Results on four video captioning benchmarks.}
    Benchmark metrics retain their original units.
    The best score among open-source models and Reviso in each column is highlighted in \colorbox{revisogreen}{green}.
    Gray closed-source rows are for reference and excluded from ranking.}
    \label{tab:argus_main}
    \begingroup
    \footnotesize
    \setlength{\tabcolsep}{2pt}
    \renewcommand{\arraystretch}{1.1}
    \begin{tabular}{
      >{\raggedright\arraybackslash}m{\dimexpr0.24\textwidth-2\tabcolsep\relax}
      *{6}{>{\centering\arraybackslash}m{\dimexpr0.19\textwidth/3-2\tabcolsep\relax}}
      *{4}{>{\centering\arraybackslash}m{\dimexpr0.095\textwidth-2\tabcolsep\relax}}
    }
    \toprule
    \multirow{2}{*}{\textbf{Model}}
    & \multicolumn{3}{c}{\textbf{DREAM-1K (\%)}}
    & \multicolumn{3}{c}{\textbf{MCTS-VCB (\%)}}
    & \multicolumn{2}{c}{\textbf{CaReBench (\%)}}
    & \multicolumn{2}{c}{\textbf{VDC (Detailed)}} \\
    \cmidrule(lr){2-4}\cmidrule(lr){5-7}\cmidrule(lr){8-9}\cmidrule(lr){10-11}
    & \textbf{Prec.}$\uparrow$ & \textbf{Rec.}$\uparrow$ & \textbf{F1}$\uparrow$
    & \textbf{Prec.}$\uparrow$ & \textbf{Rec.}$\uparrow$ & \textbf{F1}$\uparrow$
    & \textbf{Action F1}$\uparrow$ & \textbf{Object F1}$\uparrow$
    & \textbf{TP score}$\uparrow$ & \textbf{TP acc.}$\uparrow$ \\
    \midrule
    \rowcolor{revisoblue}
    \multicolumn{11}{c}{\textbf{Proprietary Models}} \\
    {\hypersetup{citecolor=gray}\graytext{GPT-6 Astra~\cite{openai2026gpt6astra}}} & \graytext{24.7} & \graytext{61.8} & \graytext{35.2} &
    \graytext{30.7} & \graytext{64.5} & \graytext{41.6} & \graytext{44.3} & \graytext{34.5} &
    \graytext{3.00} & \graytext{0.576} \\
    {\hypersetup{citecolor=gray}\graytext{Gemini 3.8 Flash~\cite{deepmind2026gemini38flash}}} & \graytext{34.8} & \graytext{50.2} & \graytext{41.1} &
    \graytext{47.8} & \graytext{50.7} & \graytext{49.2} & \graytext{39.7} & \graytext{34.3} &
    \graytext{2.46} & \graytext{0.486} \\
    \midrule
    \rowcolor{revisopink}
    \multicolumn{11}{c}{\textbf{Open-Source Models}} \\
    ARC-Qwen-Video-7B~\cite{ge2025archunyuan} & 31.2 & 31.4 & 31.3 & 58.4 & 42.7 & 49.3 & 24.9 & 30.2 & 2.90 & 0.554 \\
    InternVL3.5-8B~\cite{wang2025internvl3} & 41.2 & 18.8 & 25.8 & 61.7 & 24.0 & 34.6 & 13.2 & 19.7 & 2.12 & 0.388 \\
    InternVL3.5-38B~\cite{wang2025internvl3} & 36.0 & 34.7 & 35.3 & 65.7 & 25.6 & 36.8 & 14.3 & 20.5 & 2.12 & 0.393 \\
    LLaVA-OneVision-2~\cite{an2026llavaonevision2} & 34.7 & 24.9 & 29.0 & 64.7 & 35.6 & 45.9 & 23.5& 30.4 & 2.65 & 0.505 \\
    Molmo2-8B~\cite{clark2026molmo2} & 31.2 & 37.6 & 34.1 & 47.3 & 45.9 & 46.6 & 29.0 & 33.7 & 2.67 & 0.503 \\
    Tarsier2-7B~\cite{yuan2025tarsier2} & 41.50 & 37.1 & 39.2 & 56.5 & 35.2 & 43.4 & 30.2 & 25.4 & 2.65 & 0.505 \\
    Qwen3-VL-8B-Instruct~\cite{bai2025qwen3vl} & 30.9 & \cellcolor{revisogreen}\textbf{38.3} & 34.2 & 65.1 & 39.9 & 49.5 & 29.5 & 33.0 & 2.65 & 0.498 \\
    Qwen3-VL-32B-Instruct~\cite{bai2025qwen3vl} & 30.5 & 37.7 & 33.7 & 46.7 & 56.5 & 51.1 & 32.2 & 33.4 & 2.90 & 0.553 \\
    Qwen3-VL-32B-Thinking~\cite{bai2025qwen3vl} & 30.7 & 35.2 & 32.8 & 44.0 & 56.7  &49.5 & 33.3 & 33.6 & 2.88 & 0.547 \\
    Qwen3.5-9B~\cite{qwen2026qwen35} & 36.3 & 34.0 & 35.1 & 57.6 & 37.7 & 45.5 & 25.2 & 27.2 & 2.33 & 0.437 \\
    Qwen3.6-27B~\cite{team2026qwen327b} & 36.7 & 36.9 & 36.8 & 60.5 & 42.5 & 49.9 & 25.5 & 27.2 & 2.67 & 0.506 \\
    Qwen3.6-35B-A3B~\cite{team2026qwen3} & 38.1 & 33.2 & 35.5 & 59.7 & 41.4 & 48.9 & 24.0 & 19.7 & 2.66 & 0.504 \\
    Qwen3.8-27B~\cite{qwen2026qwen38} & 34.5 & 37.2 & 35.8 & 47.6 & 47.7 & 47.6 & 29.2 & 32.0 & 2.63 & 0.497 \\
    \midrule
    \rowcolor{revisoyellow}
    \multicolumn{11}{c}{\textbf{Our Model}} \\
    \textbf{Reviso (Ours)}$^{*}$ & \cellcolor{revisogreen}\textbf{42.3} & 37.0 & \cellcolor{revisogreen}\textbf{39.5} & \cellcolor{revisogreen}\textbf{65.7} & \cellcolor{revisogreen}\textbf{56.8} & \cellcolor{revisogreen}\textbf{60.9} & \cellcolor{revisogreen}\textbf{34.3} & \cellcolor{revisogreen}\textbf{33.9} & \cellcolor{revisogreen}\textbf{2.95} & \cellcolor{revisogreen}\textbf{0.568} \\
    \bottomrule
    \end{tabular}%
    \endgroup
    \end{table}

As shown in Table~\ref{tab:argus_main}, Reviso achieves the best overall results among the evaluated open-source models across four video captioning benchmarks, consistently outperforming its Qwen3.5-9B backbone.
On DREAM-1K, Reviso achieves the highest precision and improves F1 by 4.4 percentage points over the backbone.
More substantial gains are observed on MCTS-VCB, where precision and recall improve by 8.1 and 19.1 points, respectively, resulting in a 15.4-point F1 improvement.
On CaReBench, Reviso achieves the highest action and object F1 scores, outperforming its backbone by 9.1 and 6.7 points, respectively.
It also achieves the best results among the evaluated open-source models on both VDC Detailed metrics.
Together, these results demonstrate the effectiveness of Reviso in improving factual coverage and fine-grained video descriptions across diverse benchmarks.

\wrapneedspace{25\baselineskip}

\subsection{Ablation Study}
\label{sec:ablations}

\begin{wraptable}{r}{\smallfloatwidth}
  \centering
  \caption{\textbf{Ablation of tool use and training.}
  All scores are percentages. Base denotes Qwen3.5-9B.
  Base and full-model results follow Table~\ref{tab:argus_main}.
  Checkmarks indicate active reward components.}
  \label{tab:ablations}
  \begingroup
  \footnotesize
  \setlength{\tabcolsep}{1.4pt}
  \renewcommand{\arraystretch}{1.18}
  \begin{minipage}{\linewidth}
  \centering
  \resizebox{\linewidth}{!}{%
  \begin{tabular}{@{}lccc|ccc@{}}
    \toprule
    \multirow{2}{*}{\textbf{Variant}}
    & \multicolumn{3}{c|}{\textbf{Reward}}
    & \textbf{DREAM-1K}
    & \multicolumn{2}{c}{\textbf{CaReBench}} \\
    \cmidrule(lr){2-4}
    \cmidrule(lr){5-5}\cmidrule(lr){6-7}
    & \textbf{KFR}
    & \textbf{PAR}
    & \textbf{UTCP}
    & \textbf{F1}$\uparrow$
    & \textbf{Action F1}$\uparrow$
    & \textbf{Object F1}$\uparrow$ \\
    \midrule
    Base
    & -- & -- & --
    & 35.1 & 25.2 & 27.2 \\
    SFT only
    & -- & -- & --
    & 37.6 & 30.7 & 30.1 \\
    w/o Tool use
    & -- & -- & --
    & 35.3 & 24.8 & 27.4 \\
    \midrule
    \multicolumn{7}{c}{\textit{ GRPO reward variants}} \\
    w/o KFR
    & -- & $\checkmark$ & $\checkmark$
    & 38.2 & 32.9 & 32.1 \\
    w/o PAR
    & $\checkmark$ & -- & $\checkmark$
     & 37.9 & 31.3 & 29.7 \\
    w/o UTCP
    & $\checkmark$ & $\checkmark$ & --
   & 38.4 & 33.7 & 32.4 \\
    \textbf{Full}
    & $\checkmark$ & $\checkmark$ & $\checkmark$
    & \textbf{39.5} & \textbf{34.3} & \textbf{33.9} \\
    \bottomrule
  \end{tabular}%
  }
  \end{minipage}
  \endgroup
  \end{wraptable}

We conduct ablation studies on DREAM-1K and CaReBench, reporting F1 on DREAM-1K and action and object F1 on CaReBench.
We compare the base, SFT-only, and Reviso, together with a variant without tool use.
We also remove KFR, PAR, and UTCP individually to assess the contribution of each reward component.
As shown in Table~\ref{tab:ablations}, SFT improves all three metrics over the base model, with the largest gain in action F1.
The variant without active captioning remains close to the base model, supporting the importance of active observation.
The full GRPO objective further improves all three metrics over SFT and achieves the best results among the evaluated variants.
Removing KFR or UTCP lowers performance on both benchmarks, indicating that factual supervision and the tool-call penalty both contribute to caption quality.
Removing PAR  causes the largest reduction among the reward ablations, including a 4.2-point drop in object F1.
These results show that active observation and the full reward objective improve Reviso's ability to describe video content accurately.

\finishwrap
  \wrapneedspace{24\baselineskip}

\subsection{Analysis and Findings}

\subsubsection{Caption-To-QA Evaluation}
\label{sec:downstream_qa}

\begin{wraptable}{r}{\smallfloatwidth}
\centering
\caption{\textbf{Caption-to-QA results on two video understanding
  benchmarks.}
  We report accuracy (\%). The best result in each column is shown in bold.}
\label{tab:downstream_qa}
\begingroup
\small
\setlength{\tabcolsep}{2.5pt}
\renewcommand{\arraystretch}{1}
\resizebox{\linewidth}{!}{%
\begin{tabular}{@{}lccccc@{}}
\toprule
\multirow{2}{*}{\textbf{Model}}
& \multirow{2}{*}{\textbf{Params.}}
& \multirow{2}{*}{\textbf{LVBench}}
& \multicolumn{3}{c}{\textbf{MMVU}} \\
\cmidrule(lr){4-6}
& & & \textbf{MCQ} & \textbf{Open} & \textbf{Overall} \\
\midrule
Gemini 3.1 Pro~\cite{deepmind2026gemini31pro} & - & 41.2 & 77.6 & 46.7 & 66.1 \\
ARC-Qwen-Video~\cite{ge2025archunyuan} & 7B & 48.1 & 77.6 & 48.8 & 66.8 \\
InternVL3.5~\cite{wang2025internvl3} & 38B & 41.8 & 69.2 & 43.7 & 59.6 \\
Tarsier2~\cite{yuan2025tarsier2} & 7B & 45.8 & 76.2 & 54.2 & 67.9  \\
Qwen3.6~\cite{team2026qwen3} & 35B & 45.3 & 77.1 & 47.2 & 65.9 \\
Qwen3.8~\cite{qwen2026qwen38} & 27B & 44.5 & 81.3 & 50.1 & 69.6 \\
Qwen3.5~\cite{qwen2026qwen35} & 9B & 47.1 & 79.2 & 56.1 & 70.6 \\
\midrule
\textbf{Reviso (Ours)}$^{\dagger}$ & 9B & \textbf{52.0} & \textbf{81.7} & \textbf{57.9} & \textbf{72.7} \\
\bottomrule
\end{tabular}%
}
\endgroup
\end{wraptable}

To assess the utility of generated captions for downstream video understanding, we evaluate caption-based question answering on LVBench~\cite{wang2025lvbench} and MMVU~\cite{zhao2025mmvu}.
We use Gemini 3.1 Pro~\cite{deepmind2026gemini31pro} to extract answers to the benchmark questions from the generated captions, without access to the original videos.
All captions are generated independently of the questions.
As shown in Table~\ref{tab:downstream_qa}, Reviso achieves the highest accuracy across all four evaluation metrics.
Compared with its Qwen3.5-9B backbone, Reviso improves accuracy by 4.9 percentage points on LVBench and 2.1 points on MMVU Overall.
These results suggest that Reviso's captions better preserve information useful for downstream video understanding.

\finishwrap

\wrapneedspace{28\baselineskip}

\subsubsection{Benefits of Tool-Assisted Observation}
\label{sec:tool_observation}

\begin{figure}[!t]
\centering
\includegraphics[width=\textwidth]{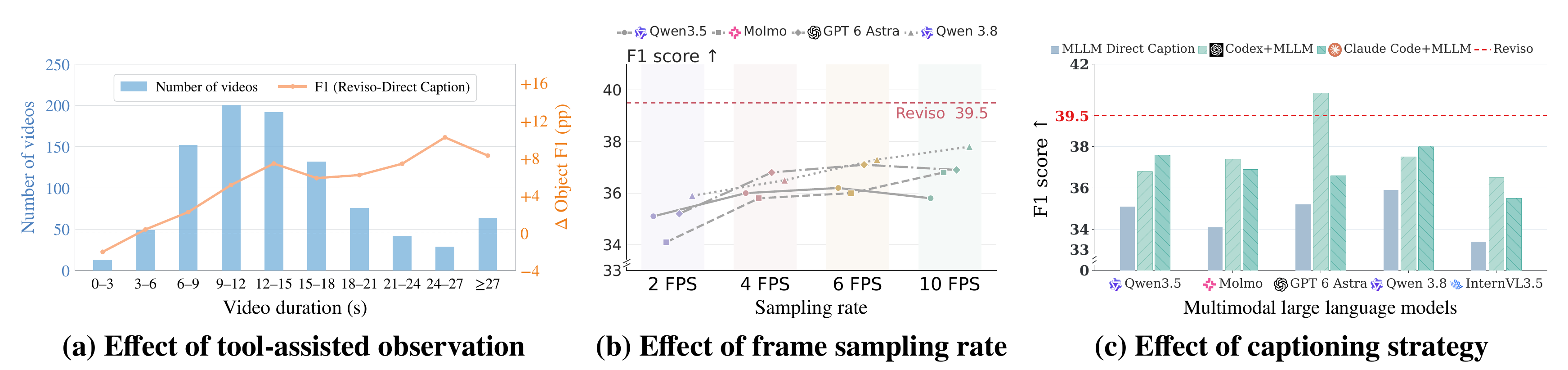}
\caption{\textbf{(a) Effect of tool-assisted observation on CaReBench.}
Bars show the number of videos in each duration group (left axis).
The line shows the object F1 difference between tool-assisted and direct captioning with the same
Reviso model (right axis, percentage points).
\textbf{(b) Effect of frame sampling rate on DREAM-1K.}
Direct captioning is evaluated at 2, 4, 6, and 10 FPS.
\textbf{(c) Effect of captioning strategy on DREAM-1K.}
Direct captioning at 2 FPS is compared with the Codex and Claude Code workflows for five VLMs.
The dashed lines in (b) and (c) mark Reviso's F1 score of 39.5.}
\label{fig:reviso_analysis}
\end{figure}

To assess the benefits of fine-grained observation, we compare Reviso with a variant that generates captions without tool assistance on CaReBench.
Figure~\ref{fig:reviso_analysis}(a) shows the difference in object F1 ($\Delta$ Object F1) between the two settings across video duration groups.
Tool-assisted observation yields larger improvements for longer videos, reaching approximately 10 percentage points for clips lasting 24--27 seconds.
For shorter videos, the additional gains are relatively limited, potentially because the frame budget already provides dense temporal coverage.
These results suggest that targeted observation is particularly useful for longer videos, where sparse sampling may overlook important visual information.

\FloatBarrier
\wrapneedspace{8\baselineskip}

\subsubsection{Does Higher FPS Improve Caption Quality}
\label{sec:fps}

To investigate the effect of frame sampling density on caption quality, we evaluate direct-captioning models with uniform sampling at different frame rates on DREAM-1K~\cite{wang2024tarsier}.
As shown in Figure~\ref{fig:reviso_analysis}(b), increasing the sampling rate generally improves caption quality, but the gains are not consistent across models.
While Molmo2-8B~\cite{clark2026molmo2} and Qwen3.8-27B~\cite{qwen2026qwen38} continue to benefit from denser sampling, Qwen3.5-9B~\cite{qwen2026qwen35} and GPT-6 Astra~\cite{openai2026gpt6astra} exhibit performance declines at 10 FPS.
Notably, Reviso outperforms all evaluated direct-captioning configurations, despite their increased sampling rates.
These results suggest that uniformly increasing frame density alone is insufficient to ensure better caption quality, highlighting the potential benefits of goal-directed visual observation.

\begin{figure}[t]
\centering
\includegraphics[width=\textwidth]{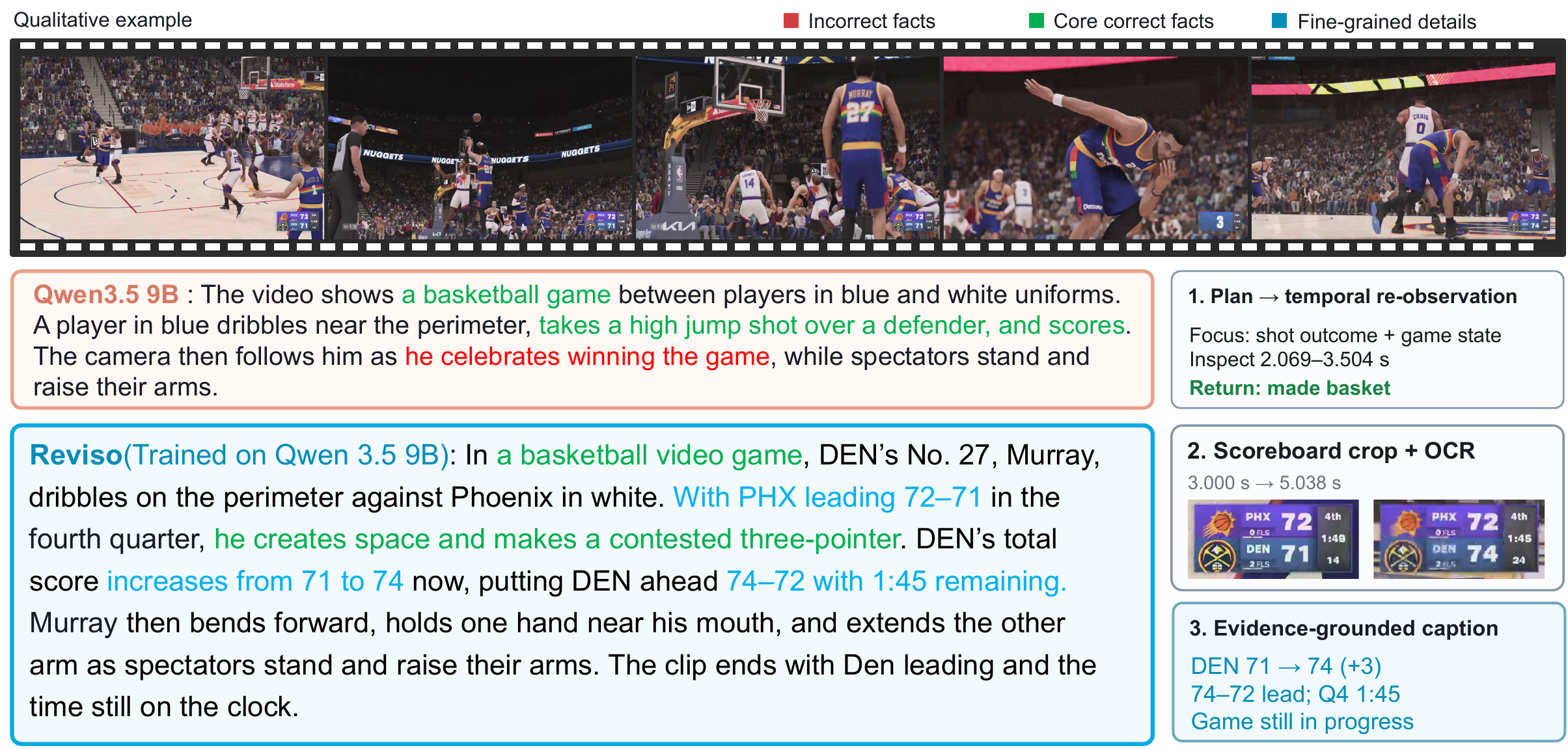}
\caption{\textbf{Qualitative comparison of Qwen3.5-9B and Reviso.}
  Reviso identifies key information to inspect and uses OCR to track the game
  state, producing a more accurate caption.
  Red, green, and blue indicate incorrect facts, key facts, and fine-grained
  details, respectively.}
\label{fig:case_studies}
\end{figure}

\subsubsection{Comparison of Captioning Strategies}
\label{sec:agentic_comparison}

To evaluate different captioning strategies, we compare Reviso with direct captioning and general-purpose agentic workflows based on Codex~\cite{openai2025codex} and Claude Code~\cite{anthropic2025claudecode}.
For direct captioning, each MLLM generates captions from uniformly sampled video frames, while the agentic workflows enable the same models to plan observations and invoke tools before generating captions.
As shown in Figure~\ref{fig:reviso_analysis}(c), agentic workflows improve caption quality for several models, although the gains vary across models and harnesses.
Notably, Reviso outperforms most evaluated model-harness combinations, with only GPT-6 Astra~\cite{openai2026gpt6astra} paired with Codex achieving a higher F1 score.
These results suggest that access to observation tools alone does not guarantee high-quality captions, highlighting the potential benefits of learning video-specific observation strategies.

\wrapneedspace{12\baselineskip}

\section{Case Study}
\label{sec:qualitative}

As illustrated in Figure~\ref{fig:case_studies}, Qwen3.5-9B correctly identifies the shooting action but mistakenly interprets the player's subsequent gesture as a celebration of victory.
In contrast, Reviso revisits the shot sequence to verify the outcome and applies cropping and OCR to extract the score and remaining time from the scoreboard.
The acquired evidence confirms that Denver takes a 74--72 lead after the three-pointer, with 1:45 remaining in the fourth quarter.
By incorporating this evidence, Reviso accurately describes the game state without incorrectly concluding that the game has ended.
This example highlights Reviso's ability to identify key information for verification, gather relevant evidence through targeted tool use, and avoid unsupported conclusions.

\section{Conclusion}
\label{sec:conclusion}

In this work, we present \textbf{Reviso}, an agentic video captioning framework that learns to prioritize important visual information and actively gather supporting evidence.
Reviso employs a hierarchical Context--Focus--Points (CFP) observation plan to guide multi-round, tool-assisted observation, adapting its plan based on newly acquired evidence.
To learn this observation strategy, we develop a verification-guided trajectory construction pipeline and introduce tailored rewards for GRPO to improve caption quality, observation planning, and regulate tool-use costs.
Experiments across four video captioning benchmarks show that Reviso outperforms the evaluated open-source baselines, while its generated captions better support downstream video question answering.
Further analyses highlight the benefits of observation planning and targeted evidence gathering for improving caption quality.
Together, these findings demonstrate the effectiveness and potential of priority-guided active captioning, where models learn what deserves attention and actively gather evidence to support accurate video descriptions.

\FloatBarrier
\bibliographystyle{reviso_refs}
\bibliography{ref}
\end{document}